\documentclass[11pt,a4paper]{article}
\usepackage[hyperref]{naaclhlt2019}
\usepackage{times}
\usepackage{latexsym}
\usepackage{graphicx}
\usepackage{todonotes}
\usepackage{soul}
\usepackage{amsmath}
\usepackage{comment}
\usepackage[skip=1ex]{caption}
\usepackage{subcaption}
\usepackage{multirow}
\graphicspath{ {figures/} }

\usepackage{url}
\usepackage{enumitem}

\setlist[enumerate]{nosep, topsep=3pt}
\setlist[itemize]{nosep,topsep=3pt}

\aclfinalcopy 

\newcommand{\Note}[2]{} 
\newcommand{\SideNote}[2]{} 
\renewcommand{\Note}[2]{\todo[color=#1,size=\small, inline=true]{#2}} 
\renewcommand{\SideNote}[2]{\todo[color=#1,size=\small]{#2}} %

\title{Better Automatic Evaluation of Open-Domain Dialogue Systems with Contextualized Embeddings}

  \author{Sarik Ghazarian \\
  Information Sciences Institute \\
  University of Southern California \\
  {\tt sarik@isi.edu} \\\And
  Johnny Tian-Zheng Wei \\
  College of Natural Sciences \\
  University of Massachusetts Amherst\\
  {\tt jwei@umass.edu} \\\AND
  Aram Galstyan \\
  Information Sciences Institute \\
  University of Southern California \\
  {\tt galstyan@isi.edu} \\\And Nanyun Peng\\
  Information Sciences Institute \\
  University of Southern California \\
  {\tt npeng@isi.edu}}

\date{}

\begin{document}
\maketitle
\begin{abstract}
  Despite advances in open-domain dialogue systems, automatic evaluation of such systems is still a challenging problem. Traditional reference-based metrics such as BLEU are ineffective because there could be many valid responses for a given context that share no common words with reference responses. 
  A recent work proposed Referenced metric and Unreferenced metric Blended Evaluation Routine (RUBER) to combine a learning-based metric, which predicts relatedness between a generated response and a given query, with reference-based metric; it showed high correlation with human judgments.
  In this paper, we explore using contextualized word embeddings to compute more accurate relatedness scores, thus better evaluation metrics. 
  Experiments show that our evaluation metrics outperform RUBER, which is trained on static embeddings. 

\end{abstract}

\section{Introduction}
Recent advances in open-domain dialogue systems (i.e. chatbots) highlight the difficulties in automatically evaluating them. This kind of evaluation inherits a characteristic challenge of NLG evaluation - given a context, there might be a diverse range of acceptable responses \cite{DBLP:journals/jair/GattK18}. 

Metrics based on $n$-gram overlaps such as BLEU \cite{DBLP:conf/acl/PapineniRWZ02} and ROUGE \cite{rouge-a-package-for-automatic-evaluation-of-summaries}, originally designed for evaluating machine translation and summarization, have been adopted to evaluate dialogue systems~\cite{DBLP:conf/naacl/SordoniGABJMNGD15, DBLP:conf/naacl/LiGBGD16, DBLP:conf/aaai/SuSHLC18}.
However, \citet{DBLP:conf/emnlp/LiuLSNCP16} found a weak segment-level correlation between these metrics and human judgments of response quality. As shown in Table \ref{qual-analysis}, high-quality responses can have low or even no $n$-gram overlap with a reference response, showing that these metrics are not suitable for dialogue evaluation \cite{DBLP:conf/emnlp/NovikovaDCR17, DBLP:conf/acl/LoweNSABP17}. 

\begin{table}[t]
\centering
\begin{tabular}{p{80mm}}
\hline \hline
\textbf{Dialogue Context} \\
Speaker 1: Hey! What are you doing here? \\
Speaker 2: I'm just shopping.\\\hline \hline
\textbf{Query}: What are you shopping for?\\ \hline
\textbf{Generated Response}: Some new clothes.\\ \hline
\textbf{Reference Response}: I want buy gift for my mom!\\ \hline
\end{tabular}
\caption{An example of zero BLEU score for an acceptable generated response in multi-turn dialogue system}
\label{qual-analysis}
\end{table}

Due to the lack of strong automatic evaluation metrics, many researchers resort primarily to human evaluation for assessing their dialogue systems performances~\cite{DBLP:conf/acl/ShangLL15, DBLP:conf/naacl/SordoniGABJMNGD15, DBLP:conf/emnlp/ShaoGBGSK17}.  There are two main problems with human annotation: 1) it is time-consuming and expensive, and 2) it does not facilitate comparisons across research papers. For certain research questions that involve hyper-parameter tuning or architecture searches, the amount of human annotation makes such studies infeasible \cite{D17-1151,melis2018on}. 
Therefore, developing reliable automatic evaluation metrics for open-domain dialog systems is imperative.

\begin{figure*}[t]
    \centering
    \includegraphics[width=12cm, height = 6.5cm]{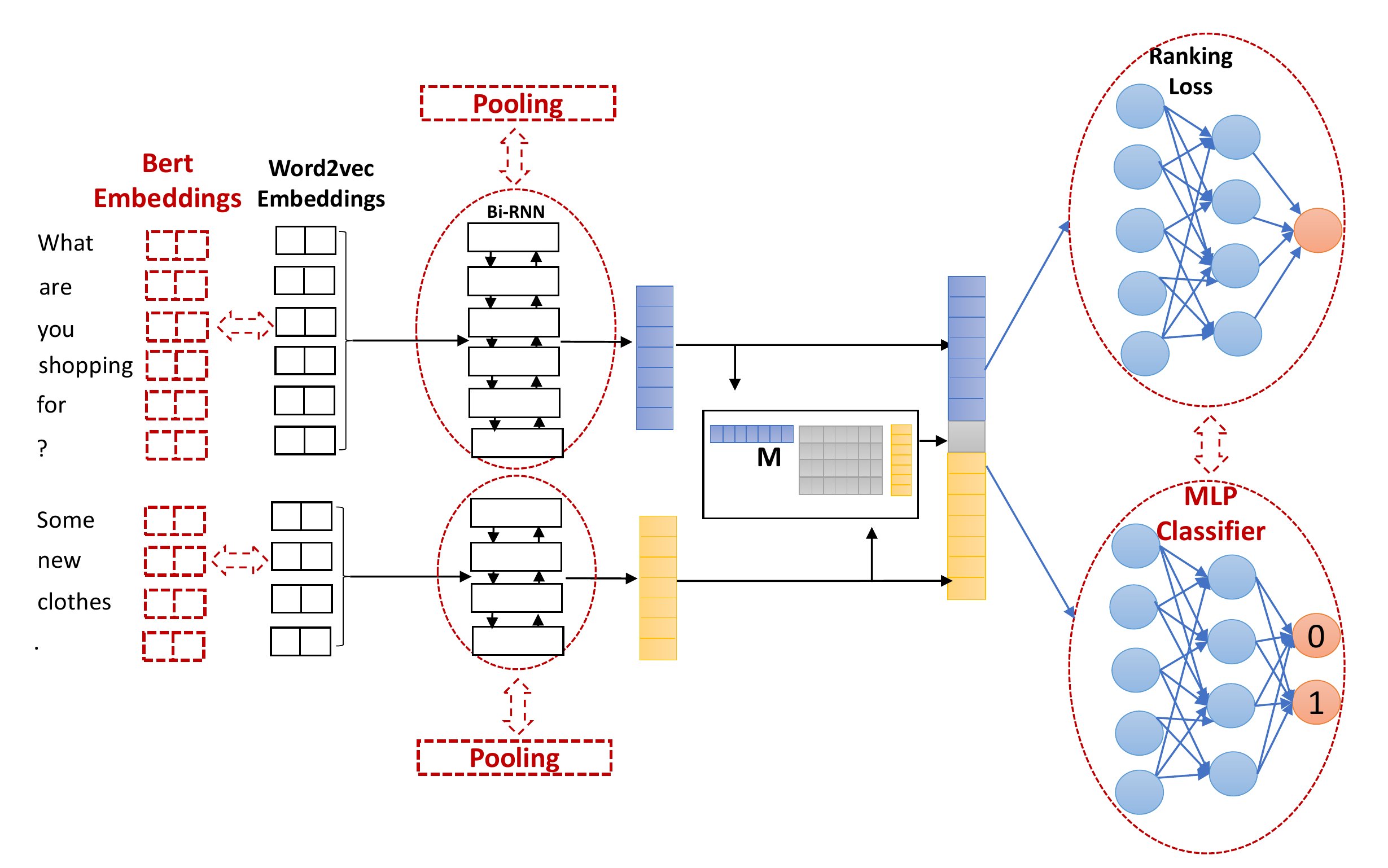}
    \caption{An illustration of changes applied to RUBER's unreferenced metric's architecture. Red dotted double arrows show three main changes. The leftmost section is related to substituting word2vec embeddings with BERT embeddings. The middle section replaces Bi-RNNs with simple pooling strategies to get sentence representations. The rightmost section switches ranking loss function to MLP classifier with cross entropy loss function.}

    \label{fig:unref}
\end{figure*}
The Referenced metric and Unreferenced metric Blended Evaluation Routine (RUBER)~\cite{DBLP:conf/aaai/TaoMZY18} stands out from recent work in automatic dialogue evaluation, relying minimally on human-annotated datasets of response quality for training. RUBER evaluates responses with a blending of scores from two metrics: 
\begin{itemize}
\item an \emph{Unreferenced} metric, which computes the relevancy of a response to a given query inspired by~\newcite{grice1975logic}'s theory that the quality of a response is determined by its relatedness and appropriateness, among other properties. This model is trained with negative sampling.
\item a \emph{Referenced} metric, which determines the similarities between generated and reference responses using word embeddings.
\end{itemize}
Both metrics strongly depend on learned word embeddings. We propose to explore the use of contextualized embeddings, specifically BERT embeddings \cite{DBLP:journals/corr/abs-1810-04805}, in composing evaluation metrics. Our contributions in this work are as follows:

\begin{itemize}
\item We explore the efficiency of contextualized word embeddings on training unreferenced models for open-domain dialog system evaluation. 
\item We explore different network architectures and objective functions to better utilize contextualized word embeddings, and show their positive effects. 
\end{itemize}

\section{Proposed models}

We conduct the research under the RUBER metric's referenced and unreferenced framework, where we replace their static word embeddings with pretrained BERT contextualized embeddings and compare the performances. We identify three points of variation with two options each in the unreferenced component of RUBER. The main changes are in the word embeddings, sentence representation, and training objectives that will be explained with details in the following section.  Our experiment follows a 2x2x2 factorial design.

\subsection{Unreferenced Metric}
The unreferenced metric predicts how much a generated response is related to a given query. Figure~\ref{fig:unref} presents RUBER's unreferenced metric overlaid with our proposed changes in three parts of the architecture. Changes are illustrated by red dotted double arrows and include word embeddings, sentence representation and the loss function.

\subsubsection{Word Embeddings}
Static and contextualized embeddings are two different types of word embeddings that we explored.
\begin{itemize}
\item {\bf Word2vec.} Recent works on learnable evaluation metrics use simple word embeddings such as word2vec and GLoVe as input to their models \cite{DBLP:conf/aaai/TaoMZY18, DBLP:conf/acl/LoweNSABP17, DBLP:journals/corr/KannanV17}. Since these static embeddings have a fixed context-independent representation for each word, they cannot represent the rich semantics of words in contexts. 

\item {\bf BERT.} Contextualized word embeddings are recently shown to be beneficial in many NLP tasks \cite{DBLP:journals/corr/abs-1810-04805, radford2018improving, DBLP:conf/naacl/PetersNIGCLZ18, DBLP:journals/corr/abs-1903-08855}. A noticeable contextualized word embeddings, BERT~\cite{DBLP:journals/corr/abs-1810-04805}, is shown to perform competitively among other contextualized embeddings, thus we explore the effect of BERT embeddings on open domain dialogue systems evaluation task. Specifically, we substitute the word2vec embeddings with BERT embeddings in RUBER's unreferenced score as shown in the leftmost section of Figure~\ref{fig:unref}. 
\end{itemize}

\subsubsection{Sentence Representation}
This section composes a single vector representation for both a query and a response.
\begin{itemize}
    \item {\bf Bi-RNN.} In the RUBER model, Bidirectional Recurrent Neural Networks (Bi-RNNs) are trained for this purpose. 
    
    \item {\bf Pooling.} We explore the effect of replacing Bi-RNNs with some simple pooling strategies on top of words BERT embeddings (middle dotted section in Figure~\ref{fig:unref}). The intuition behind this is that BERT embeddings are pre-trained on bidirectional transformers and they include complete information about word's context, therefore, another layer of bi-RNNs could just blow up the number of parameters with no real gains.
\end{itemize}

\subsubsection{MLP Network}
Multilayer Perceptron Network (MLP) is the last section of RUBER's unreferenced model that is trained by applying negative sampling technique to add some random responses for each query into training dataset. 
\begin{itemize}
    \item {\bf Ranking loss.} The objective is to maximize the difference between relatedness score predicted for positive and randomly added pairs. We refer to this objective function as a ranking loss function. The sigmoid function used in the last layer of MLP assigns a score to each pair of query and response, which indicates how much the response is related to a given query.
    \item {\bf Cross entropy loss.} We explore the efficiency of using a simpler loss function such as cross entropy. In fact, we consider unreferenced score prediction as a binary classification problem and replace baseline trained MLP with MLP classifier (right dotted section in Figure~\ref{fig:unref}). Since we do not have a human labeled dataset, we use negative sampling strategy to add randomly selected responses to queries in training dataset. We assign label 1 to original pairs of queries and responses and 0 to the negative samples. The output of softmax function in the last layer of MLP classifier indicates the relatedness score for each pair of query and response.
\end{itemize}

\begin{figure}[!t]
    \centering
    \includegraphics[width=7cm, height=5cm]{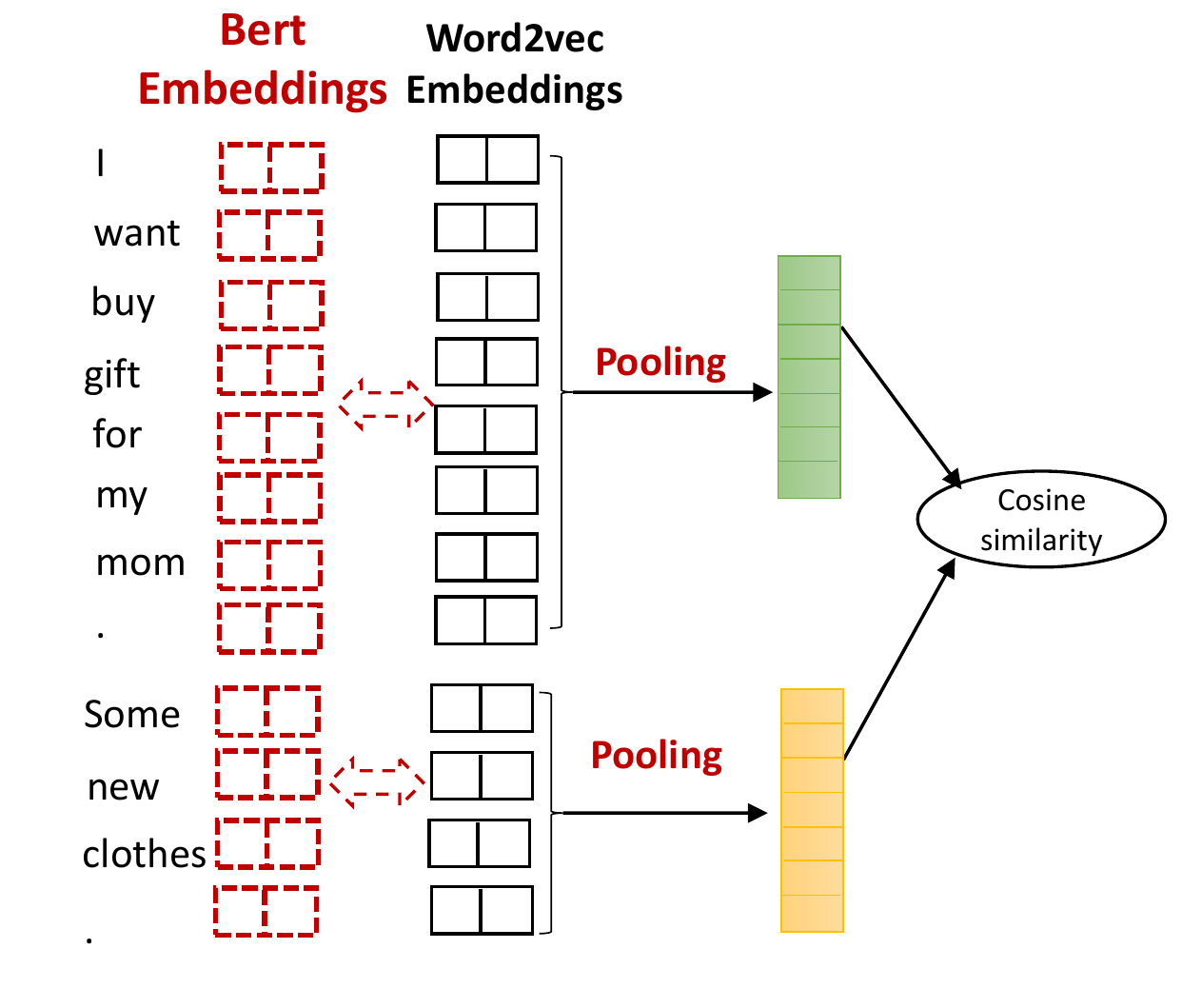}
    \caption{BERT-based referenced metric. Static word2vec embeddings are replaced with BERT embeddings (red dotted section).}
    \label{fig:ref}
\end{figure}

\begin{table*}[th!]
\centering
\begin{tabular}{l l c}
\hline \hline \bf Query & \bf  Response  & \bf Human rating \\
\hline \hline
Can I try this one on? & Yes, of course. & 5, 5, 5 \\ \hline
This is the Bell Captain's Desk. May I help you? & No, it was nothing to leave. &  1, 2, 1 \\ \hline
\begin{minipage}[t]{0.9\columnwidth}%
Do you have some experiences to share with me? I want to have a try. %
\end{minipage} 
 & 
\begin{minipage}[t]{0.6\columnwidth}%
Actually, it good to say. Thanks a lot. %
\end{minipage} & 3, 2, 2 \\ \hline
\end{tabular}
\caption{Examples of query-response pairs, each rated by three AMT workers with scores from 1 (not appropriate response) to 5 (completely appropriate response).}
\label{rating-examples}
\end{table*}

\subsection{Referenced Metric}
The referenced metric computes the similarity between generated and reference responses. RUBER achieves this by applying pooling strategies on static word embeddings to get sentence embeddings for both generated and reference responses. 
In our metric, we replace the word2vec embeddings with BERT embeddings (red dotted section in Figure~\ref{fig:ref}) to explore the effect of contextualized embeddings on calculating the referenced score. We refer to this metric as BERT-based referenced metric.

\section{Dataset} 
We used the DailyDialog dataset\footnote{\url{http://yanran.li/dailydialog}} which contains high quality multi-turn conversations about daily life including various topics~\cite{DBLP:conf/ijcnlp/LiSSLCN17}, to train our dialogue system as well as the evaluation metrics. This dataset includes almost 13k multi-turn dialogues between two parties splitted into 42,000/3,700/3,900 query-response pairs for train/test/validation sets. We divided these sets into two parts, the first part for training dialogue system and the second part for training unreferneced metric.

\subsection{Generated responses} 
We used the first part of train/test/validation sets with overall 20,000/1,900/1,800 query-response pairs to train an attention-based sequence-to-sequence (seq2seq) model \cite{DBLP:journals/corr/BahdanauCB14} and generate responses for evaluation. We used OpenNMT \cite{opennmt} toolkit to train the model. The encoder and decoder are Bi-LSTMs with 2 layers each containing 500-dimensional hidden units. We used 300-dimensional pretrained word2vec embeddings as our word embeddings. The model was trained by using SGD optimizer with learning rate of 1. We used random sample with temperature control and set temperature value to 0.01 empirically to get grammatical and diverse responses. 

\subsection{Human Judgments}
We collected human annotations on generated responses  in order to  compute  the  correlation  between  human judgments and automatic evaluation metrics.
Human annotations were collected from Amazon Mechanical Turk (AMT). AMT workers were provided a set of query-response pairs and asked to rate each pair based on the appropriateness of the response for the given query on a scale of 1-5 (not appropriate to very appropriate). 
Each survey included 5 query-response pairs with an extra pair for attention checking. We removed all pairs that were rated by workers who failed to correctly answer attention-check tests. Each pair was annotated by 3 individual turkers. Table~\ref{rating-examples} demonstrates three query-response pairs rated by three AMT workers. In total 300 utterance pairs were rated from contributions of 106 unique workers.

\section{Experimental Setup}

\subsection{Static Embeddings}
To compare how the word embeddings affect the evaluation metric, which is the main focus of this paper, we used word2vec as static embedddings trained on about 100 billion words of Google News Corpus. These 300 dimensional word embeddings include almost 3 million words and phrases. We applied these pretrained embeddings as input to dialogue generation, referenced and unreferenced metrics. 

\subsection{Contextualized Embeddings}
In order to explore the effects of contextualized embedding on evaluation metrics, we used the BERT base model with 768 vector dimensions pretrained on Books Corpus and English Wikipedia with 3,300M words \cite{DBLP:journals/corr/abs-1810-04805}. 

\subsection{Training Unreferenced model}
We used the second part of the DailyDialog dataset composed of 22,000/1,800/2,100 train/test/validation pairs to train and tune the unreferenced model, which is implemented with Tensorflow. For sentence encoder, we used 2 layers of bidirectional gated recurrent unit (Bi-GRU) with 128-dimensional hidden unit. We used three layers for MLP with 256, 512 and 128-dimensional hidden units and tanh as activation function for computing both ranking loss and cross-entropy loss. We used Adam \cite{DBLP:journals/corr/KingmaB14} optimizer with initial learning rate of $10^{-4}$ and applied learning rate decay when no improvement was observed on validation data for five consecutive epochs. We applied early stop mechanism and stopped training process after observing 20 epochs with no reduction in loss value.

\begin{table*}[t]
\begin{center}
\begin{tabular}{c|c|c|c|c|c}
\hline \hline \bf Embedding & \bf Representation & \bf Objective & \bf Pearson \tiny (p-value) & \bf Spearman \tiny (p-value) & \bf \begin{minipage}{0.2\columnwidth}%
\vspace{1mm}
\hspace{1mm}
Cosine \\Similarity
\end{minipage} \\ \hline \hline
\multirow{6}{*}{word2vec} & \multirow{2}{*}{Bi-RNN} & Ranking & 0.28 \tiny(\textless6e-7)  & 0.30 \tiny(\textless8e-8)  & 0.56  \\
\cline{3-6}
&  & Cross-Entropy & 0.22 \tiny(\textless9e-5) & 0.25 \tiny(\textless9e-6) & 0.53 \\
\cline{2-6}
~ & \multirow{2}{*}{Max Pooling} & Ranking  &  0.19 \tiny(\textless8e-4) & 0.18\tiny(\textless1e-3) & 0.50   \\
\cline{3-6}
~ & ~ & Cross-Entropy & 0.25 \tiny(\textless2e-5) & 0.25\tiny(\textless2e-5) & 0.53\\
\cline{2-6}
~ & \multirow{2}{*}{Mean Pooling} & Ranking & 0.16 \tiny(\textless5e-3) & 0.18\tiny(\textless2e-3) & 0.50\\
\cline{3-6}
~ & ~ & Cross-Entropy & 0.04 \tiny(\textless5e-1) & 0.06\tiny(\textless3e-1) & 0.47 \\
\hline
\multirow{6}{*}{BERT} & \multirow{2}{*}{Bi-RNN} & Ranking & 0.38 \tiny(\textless1e-2) & 0.31\tiny(\textless4e-8)  & 0.60 \\
\cline{3-6}
~ & ~ & Cross-Entropy &  0.29 \tiny(\textless2e-7)  & 0.24 \tiny(\textless3e-5)  & 0.55 \\
\cline{2-6}
~ & \multirow{2}{*}{Max Pooling} & Ranking & 0.41 \tiny(\textless1e-2)  & 0.36 \tiny(\textless7e-9)  & 0.65 \\
\cline{3-6}
~ & ~ & Cross-Entropy & {\bf 0.55} \tiny(\textless1e-2)  & {\bf 0.45 } \tiny(\textless1e-2)  & {\bf 0.70} \\
\cline{2-6}
~ & \multirow{2}{*}{Mean Pooling} & Ranking & 0.34 \tiny(\textless2e-9)  & 0.27 \tiny(\textless2e-6)  & 0.57 \\
\cline{3-6}
~ & ~ & Cross-Entropy & 0.32 \tiny(\textless2e-8)  & 0.29 \tiny(\textless5e-7)  & 0.55 \\
\hline
\end{tabular}
\end{center}
\caption{Correlations and similarity values between relatedness scores predicted by different unreferenced models and human judgments. First row is RUBER's unreferenced model.}
\label{unref-corr}
\end{table*}

\section{Results}
 We first present the unreferenced metrics' performances. Then, we present results on the full RUBER's framework - combining unreferenced and referenced metrics. To evaluate the performance of our metrics, we calculated the Pearson and Spearman correlations between learned metric scores and human judgments on  300 query-response pairs collected from AMT. The Pearson coefficient measures a linear correlation between two ordinal variables, while the Spearman coefficient measures any monotonic relationship. The third metric we used to evaluate our metric is cosine similarity, which computes how much the scores produced by learned metrics are similar to human scores. 

\subsection{Unreferenced Metrics Results}

This section analyzes the performance of unreferenced metrics which are trained based on various word embeddings, sentence representations and objective functions. The results in the upper section of Table~\ref{unref-corr} are all based on word2vec embeddings while the lower section are based on BERT embeddings. The first row of table~\ref{unref-corr} corresponds to RUBER's unreferenced model and the five following rows are our exploration of different unreferenced models based on word2vec embeddings, for fair comparison with BERT embedding-based ones. 
Table~\ref{unref-corr} demonstrates that unreferenced metrics based on BERT embeddings have higher correlation and similarity with human scores. Contextualized embeddings have been found to carry richer information and the inclusion of these vectors in the unreferenced metric generally leads to better performance \cite{DBLP:journals/corr/abs-1903-08855}. 

Comparing different sentence encoding strategies (Bi-RNN v.s. Pooling) by keeping other variations constant, we observe that pooling of BERT embeddings yields better performance. This would be because of BERT embeddings are pretrained on deep bidirectional transformers and using pooling mechanisms is enough to assign rich representations to sentences. In contrast, the models based on word2vec embeddings benefit from Bi-RNN based sentence encoder. Across settings, max pooling always outperforms mean pooling.
Regarding the choice of objective functions, ranking loss generally performs better for models based on word2vec embeddings, while the best model with BERT embeddings is obtained by using cross-entropy loss. We consider this as an interesting observation and leave further investigation for future research.

\begin{table*}[t]
\centering
\footnotesize{
\begin{tabular}{c|c|c|c|c|c|c|c|c@{ }}
\hline
\hline
 \multirow{2}{*}{\textbf{Model}} &
\multicolumn{3}{c|}{\textbf{Unreferenced}}  & \textbf{Referenced} & \multirow{2}{*}{\textbf{Pooling}} & \multirow{2}{*}{\textbf{Pearson}} & \multirow{2}{*}{\textbf{Spearman}} & \multirow{2}{*}{\shortstack[c]{\textbf{Cosine}\\\textbf{Similarity}}} \\
\cline{2-5}
& Embedding & Representation & Objective & Embedding & & & & \\
\hline
\hline
\multirow{3}{*}{RUBER} &
\multirow{3}{*}{word2vec} & \multirow{3}{*}{Bi-RNN} & \multirow{3}{*}{Ranking} & \multirow{3}{*}{word2vec} & min & 0.08 \tiny{(\textless 0.16)} & 0.06 \tiny{(\textless 0.28)} & 0.51 \\
& & & & & max & 0.19 \tiny{(\textless 1e-3)} & 0.23 \tiny{(\textless 4e-5)} & 0.60 \\
& & & & & mean & 0.22 \tiny{(\textless 9e-5)} & 0.21 \tiny{(\textless 3e-4)} & 0.63 \\
\hline
\multirow{3}{*}{Ours} &
\multirow{3}{*}{BERT} & \multirow{3}{*}{max Pooling} & \multirow{3}{*}{\shortstack[c]{Cross-\\Entropy}} &
\multirow{3}{*}{BERT} & min & 0.05 \tiny{(\textless 0.43)} & 0.09 \tiny{(\textless 0.13)} & 0.52 \\
& & & & & max & \textbf{0.49} \tiny{(\textless 1e-2)} & \textbf{0.44} \tiny{(\textless 1e-2)} & 0.69 \\
& & & & & mean & 0.45 \tiny{(\textless 1e-2)} & 0.34 \tiny{(\textless 1e-2)} & 0.70 \\
\hline
\end{tabular}
}
\caption{Correlation and similarity values between automatic evaluation metrics (combination of Referenced and Unreferenced metrics) and human annotations for 300 query-response pairs annotated by AMT workers. The "Pooling" column shows the combination type of referenced and unreferenced metrics.}
\label{ref+unref-corr}
\end{table*}

\subsection{Unreferenced + Referenced Metrics Results}
This section analyzes the performance of integrating variants of unreferenced metrics into the full RUBER framework which is the combination of unreferenced and referenced metrics. We only considered the best unreferenced models from Table~\ref{unref-corr}. As it is shown in Table~\ref{ref+unref-corr}, across different settings, max combinations of referenced and unereferenced metrics yields the best performance. We see that metrics based on BERT embeddings have higher Pearson and Spearman correlations with human scores than RUBER (the first row of Table \ref{ref+unref-corr}) which is based on word2vec embeddings. 

In comparison with purely unreferenced metrics (Table \ref{unref-corr}), correlations decreased across the board. This suggests that the addition of the referenced component is not beneficial, contradicting RUBER's findings \cite{DBLP:conf/aaai/TaoMZY18}. We hypothesize that this could be due to data and/or language differences, and leave further investigation for future work. 

\section{Related Work} 
Due to the impressive development of open domain dialogue systems, existence of automatic evaluation metrics can be particularly desirable to easily compare the quality of several models. 

\subsection{Automatic Heuristic Evaluation Metrics}
In some group of language generation tasks such as machine translation and text summarization, $n$-grams overlapping metrics have a high correlation with human evaluation. BLEU and METEOR are primarily used for evaluating the quality of translated sentence based on computing $n$-gram precisions and harmonic mean of precision and recall, respectively \cite{DBLP:conf/acl/PapineniRWZ02, DBLP:conf/acl/BanerjeeL05}. ROUGE computes F-measure based on the longest common subsequence and is highly applicable for evaluating text summarization \cite{rouge-a-package-for-automatic-evaluation-of-summaries}. The main drawback of mentioned $n$-gram overlap metrics, which makes them inapplicable in dialogue system evaluation is that they don't consider the semantic similarity between sentences \cite{DBLP:conf/emnlp/LiuLSNCP16, DBLP:conf/emnlp/NovikovaDCR17, DBLP:conf/acl/LoweNSABP17}. These word overlapping metrics are not compatible with the nature of language generation, which allows a concept to be appeared in different sentences with no common $n$-grams, while they all share the same meaning.

\subsection{Automatic Learnable Evaluation Metrics}
Beside the heuristic metrics, researchers recently tried to develop some trainable metrics for automatically checking the quality of generated responses. \newcite{DBLP:conf/acl/LoweNSABP17} trained a hierarchical neural network model called Automatic Dialogue Evaluation Model (ADEM) to predict the appropriateness score of dialogue responses. 
 For this purpose, they collected a training dataset by asking human about the informativeness score for various responses of a given context. However, ADEM predicts highly correlated scores with human judgments in both sentence and system level, collecting human annotation by itself is an effortful and laborious task. 

\newcite{DBLP:journals/corr/KannanV17} followed the GAN model's structure and trained a discriminator that tries to discriminate the model's generated response from human responses. Even though they found discriminator can be useful for automatic evaluation systems, they mentioned that it can not completely address the evaluation challenges in dialogue systems.

RUBER is another learnable metric, which considers both relevancy and similarity concepts for evaluation process \cite{DBLP:conf/aaai/TaoMZY18}. Referenced metric of RUBER measures the similarity between vectors of generated and reference responses computed by pooling word embeddings, while unreferenced metric uses negative sampling to train the relevancy score of generated response to a given query. Despite ADEM score, which is trained on human annotated dataset, RUBER is not limited to any human annotation. In fact, training with negative samples makes RUBER to be more general. It is obvious that both referenced and unreferenced metrics are under the influence of word embeddings information. In this work, we show that contextualized embeddings that include much more information about words and their context can have good effects on the accuracy of evaluation metrics.

\subsection{Static and Contextualized Words Embeddings}
Recently, there has been significant progress in word embedding methods. Unlike previous static word embeddings like word2vec \footnote{https://code.google.com/archive/p/word2vec/}, which maps words to constant embeddings, contextualized embeddings such as ELMo, OpenAI GPT and BERT consider word embeddings as a function of the word's context in which the word is appeared \cite{DBLP:conf/nips/McCannBXS17,DBLP:conf/naacl/PetersNIGCLZ18, radford2018improving, DBLP:journals/corr/abs-1810-04805}. ELMo learns word vectors from a deep language model pretrained on a large text corpus \cite{DBLP:conf/naacl/PetersNIGCLZ18}. OpenAI GPT uses transformers to learn a language model and also to fine-tune it for specific natural language understanding tasks \cite{radford2018improving}. BERT learns words' representations by jointly conditioning on both left and right context in training all levels of deep bidirectional transformers \cite{DBLP:journals/corr/abs-1810-04805}. In this paper, we show that beside positive effects of contexualized embeddings on many NLP tasks including question answering, sentiment analysis and semantic similarity, BERT embeddings also have the potential to help evaluate open domain dialogue systems closer to what would human do.

\section{Conclusion and Future work}
In this paper, we explored applying contextualized word embeddings to automatic evaluation of open-domain dialogue systems. 
The experiments showed that the unreferenced scores of RUBER metric can be improved by considering contextualized word embeddings which include richer representations of words and their context.

In the future, we plan to extend the work to evaluate multi-turn dialogue systems, as well as adding other aspects, such as creativity and novelty into consideration in our evaluation metrics.

\section{Acknowledgments}
We thank the anonymous reviewers for their constructive feedback, as well as the members of the PLUS lab for their useful discussion and feedback. This work is supported by Contract W911NF-15- 1-0543 with the US Defense Advanced Research Projects Agency (DARPA).
\bibliography{naaclhlt2019}

\begin{thebibliography}{27}
\expandafter\ifx\csname natexlab\endcsname\relax\def\natexlab#1{#1}\fi

\bibitem[{Bahdanau et~al.(2014)Bahdanau, Cho, and
  Bengio}]{DBLP:journals/corr/BahdanauCB14}
Dzmitry Bahdanau, Kyunghyun Cho, and Yoshua Bengio. 2014.
\newblock \href {http://arxiv.org/abs/1409.0473} {Neural machine translation by
  jointly learning to align and translate}.
\newblock \emph{CoRR}, abs/1409.0473.

\bibitem[{Banerjee and Lavie(2005)}]{DBLP:conf/acl/BanerjeeL05}
Satanjeev Banerjee and Alon Lavie. 2005.
\newblock \href {https://aclanthology.info/papers/W05-0909/w05-0909} {{METEOR:}
  an automatic metric for {MT} evaluation with improved correlation with human
  judgments}.
\newblock In \emph{Proceedings of the Workshop on Intrinsic and Extrinsic
  Evaluation Measures for Machine Translation and/or Summarization@ACL 2005,
  Ann Arbor, Michigan, USA, June 29, 2005}, pages 65--72.

\bibitem[{Britz et~al.(2017)Britz, Goldie, Luong, and Le}]{D17-1151}
Denny Britz, Anna Goldie, Minh-Thang Luong, and Quoc Le. 2017.
\newblock \href {https://doi.org/10.18653/v1/D17-1151} {Massive exploration of
  neural machine translation architectures}.
\newblock In \emph{Proceedings of the 2017 Conference on Empirical Methods in
  Natural Language Processing}, pages 1442--1451. Association for Computational
  Linguistics.

\bibitem[{Devlin et~al.(2018)Devlin, Chang, Lee, and
  Toutanova}]{DBLP:journals/corr/abs-1810-04805}
Jacob Devlin, Ming{-}Wei Chang, Kenton Lee, and Kristina Toutanova. 2018.
\newblock \href {http://arxiv.org/abs/1810.04805} {{BERT:} pre-training of deep
  bidirectional transformers for language understanding}.
\newblock \emph{CoRR}, abs/1810.04805.

\bibitem[{Gatt and Krahmer(2018)}]{DBLP:journals/jair/GattK18}
Albert Gatt and Emiel Krahmer. 2018.
\newblock \href {https://doi.org/10.1613/jair.5477} {Survey of the state of the
  art in natural language generation: Core tasks, applications and evaluation}.
\newblock \emph{J. Artif. Intell. Res.}, 61:65--170.

\bibitem[{Grice(1975)}]{grice1975logic}
H.~Paul Grice. 1975.
\newblock Logic and conversation.
\newblock In Peter Cole and Jerry~L. Morgan, editors, \emph{Speech Acts},
  volume~3 of \emph{Syntax and Semantics}, pages 41--58. Academic Press, New
  York.

\bibitem[{Kannan and Vinyals(2017)}]{DBLP:journals/corr/KannanV17}
Anjuli Kannan and Oriol Vinyals. 2017.
\newblock \href {http://arxiv.org/abs/1701.08198} {Adversarial evaluation of
  dialogue models}.
\newblock \emph{CoRR}, abs/1701.08198.

\bibitem[{Kingma and Ba(2015)}]{DBLP:journals/corr/KingmaB14}
Diederik~P. Kingma and Jimmy Ba. 2015.
\newblock \href {http://arxiv.org/abs/1412.6980} {Adam: {A} method for
  stochastic optimization}.
\newblock In \emph{3rd International Conference on Learning Representations,
  {ICLR} 2015, San Diego, CA, USA, May 7-9, 2015, Conference Track
  Proceedings}.

\bibitem[{Klein et~al.(2017)Klein, Kim, Deng, Senellart, and Rush}]{opennmt}
Guillaume Klein, Yoon Kim, Yuntian Deng, Jean Senellart, and Alexander~M. Rush.
  2017.
\newblock \href {https://doi.org/10.18653/v1/P17-4012} {Open{NMT}: Open-source
  toolkit for neural machine translation}.
\newblock In \emph{Proc. ACL}.

\bibitem[{Li et~al.(2016)Li, Galley, Brockett, Gao, and
  Dolan}]{DBLP:conf/naacl/LiGBGD16}
Jiwei Li, Michel Galley, Chris Brockett, Jianfeng Gao, and Bill Dolan. 2016.
\newblock \href {http://aclweb.org/anthology/N/N16/N16-1014.pdf} {A
  diversity-promoting objective function for neural conversation models}.
\newblock In \emph{{NAACL} {HLT} 2016, The 2016 Conference of the North
  American Chapter of the Association for Computational Linguistics: Human
  Language Technologies, San Diego California, USA, June 12-17, 2016}, pages
  110--119.

\bibitem[{Li et~al.(2017)Li, Su, Shen, Li, Cao, and
  Niu}]{DBLP:conf/ijcnlp/LiSSLCN17}
Yanran Li, Hui Su, Xiaoyu Shen, Wenjie Li, Ziqiang Cao, and Shuzi Niu. 2017.
\newblock \href {https://aclanthology.info/papers/I17-1099/i17-1099}
  {Dailydialog: {A} manually labelled multi-turn dialogue dataset}.
\newblock In \emph{Proceedings of the Eighth International Joint Conference on
  Natural Language Processing, {IJCNLP} 2017, Taipei, Taiwan, November 27 -
  December 1, 2017 - Volume 1: Long Papers}, pages 986--995.

\bibitem[{Lin(2004)}]{rouge-a-package-for-automatic-evaluation-of-summaries}
Chin-Yew Lin. 2004.
\newblock \href
  {https://www.microsoft.com/en-us/research/publication/rouge-a-package-for-automatic-evaluation-of-summaries/}
  {Rouge: a package for automatic evaluation of summaries}.

\bibitem[{Liu et~al.(2016)Liu, Lowe, Serban, Noseworthy, Charlin, and
  Pineau}]{DBLP:conf/emnlp/LiuLSNCP16}
Chia{-}Wei Liu, Ryan Lowe, Iulian Serban, Michael Noseworthy, Laurent Charlin,
  and Joelle Pineau. 2016.
\newblock \href {http://aclweb.org/anthology/D/D16/D16-1230.pdf} {How {NOT} to
  evaluate your dialogue system: An empirical study of unsupervised evaluation
  metrics for dialogue response generation}.
\newblock In \emph{Proceedings of the 2016 Conference on Empirical Methods in
  Natural Language Processing, {EMNLP} 2016, Austin, Texas, USA, November 1-4,
  2016}, pages 2122--2132. The Association for Computational Linguistics.

\bibitem[{Liu et~al.(2019)Liu, Gardner, Belinkov, Peters, and
  Smith}]{DBLP:journals/corr/abs-1903-08855}
Nelson~F. Liu, Matt Gardner, Yonatan Belinkov, Matthew Peters, and Noah~A.
  Smith. 2019.
\newblock \href {http://arxiv.org/abs/1903.08855} {Linguistic knowledge and
  transferability of contextual representations}.
\newblock \emph{CoRR}, abs/1903.08855.

\bibitem[{Lowe et~al.(2017)Lowe, Noseworthy, Serban, Angelard{-}Gontier,
  Bengio, and Pineau}]{DBLP:conf/acl/LoweNSABP17}
Ryan Lowe, Michael Noseworthy, Iulian~Vlad Serban, Nicolas Angelard{-}Gontier,
  Yoshua Bengio, and Joelle Pineau. 2017.
\newblock \href {https://doi.org/10.18653/v1/P17-1103} {Towards an automatic
  turing test: Learning to evaluate dialogue responses}.
\newblock In \emph{Proceedings of the 55th Annual Meeting of the Association
  for Computational Linguistics, {ACL} 2017, Vancouver, Canada, July 30 -
  August 4, Volume 1: Long Papers}, pages 1116--1126. Association for
  Computational Linguistics.

\bibitem[{McCann et~al.(2017)McCann, Bradbury, Xiong, and
  Socher}]{DBLP:conf/nips/McCannBXS17}
Bryan McCann, James Bradbury, Caiming Xiong, and Richard Socher. 2017.
\newblock \href
  {http://papers.nips.cc/paper/7209-learned-in-translation-contextualized-word-vectors}
  {Learned in translation: Contextualized word vectors}.
\newblock In \emph{Advances in Neural Information Processing Systems 30: Annual
  Conference on Neural Information Processing Systems 2017, 4-9 December 2017,
  Long Beach, CA, {USA}}, pages 6297--6308.

\bibitem[{McIlraith and Weinberger(2018)}]{DBLP:conf/aaai/2018}
Sheila~A. McIlraith and Kilian~Q. Weinberger, editors. 2018.
\newblock \href
  {https://www.aaai.org/ocs/index.php/AAAI/AAAI18/schedConf/presentations}
  {\emph{Proceedings of the Thirty-Second {AAAI} Conference on Artificial
  Intelligence, (AAAI-18), the 30th innovative Applications of Artificial
  Intelligence (IAAI-18), and the 8th {AAAI} Symposium on Educational Advances
  in Artificial Intelligence (EAAI-18), New Orleans, Louisiana, USA, February
  2-7, 2018}}. {AAAI} Press.

\bibitem[{Melis et~al.(2018)Melis, Dyer, and Blunsom}]{melis2018on}
Gábor Melis, Chris Dyer, and Phil Blunsom. 2018.
\newblock \href {https://openreview.net/forum?id=ByJHuTgA-} {On the state of
  the art of evaluation in neural language models}.
\newblock In \emph{International Conference on Learning Representations}.

\bibitem[{Novikova et~al.(2017)Novikova, Dusek, Curry, and
  Rieser}]{DBLP:conf/emnlp/NovikovaDCR17}
Jekaterina Novikova, Ondrej Dusek, Amanda~Cercas Curry, and Verena Rieser.
  2017.
\newblock \href {https://aclanthology.info/papers/D17-1238/d17-1238} {Why we
  need new evaluation metrics for {NLG}}.
\newblock In \emph{Proceedings of the 2017 Conference on Empirical Methods in
  Natural Language Processing, {EMNLP} 2017, Copenhagen, Denmark, September
  9-11, 2017}, pages 2241--2252.

\bibitem[{Papineni et~al.(2002)Papineni, Roukos, Ward, and
  Zhu}]{DBLP:conf/acl/PapineniRWZ02}
Kishore Papineni, Salim Roukos, Todd Ward, and Wei{-}Jing Zhu. 2002.
\newblock \href {http://www.aclweb.org/anthology/P02-1040.pdf} {Bleu: a method
  for automatic evaluation of machine translation}.
\newblock In \emph{Proceedings of the 40th Annual Meeting of the Association
  for Computational Linguistics, July 6-12, 2002, Philadelphia, PA, {USA.}},
  pages 311--318. {ACL}.

\bibitem[{Peters et~al.(2018)Peters, Neumann, Iyyer, Gardner, Clark, Lee, and
  Zettlemoyer}]{DBLP:conf/naacl/PetersNIGCLZ18}
Matthew~E. Peters, Mark Neumann, Mohit Iyyer, Matt Gardner, Christopher Clark,
  Kenton Lee, and Luke Zettlemoyer. 2018.
\newblock \href {https://aclanthology.info/papers/N18-1202/n18-1202} {Deep
  contextualized word representations}.
\newblock In \emph{Proceedings of the 2018 Conference of the North American
  Chapter of the Association for Computational Linguistics: Human Language
  Technologies, {NAACL-HLT} 2018, New Orleans, Louisiana, USA, June 1-6, 2018,
  Volume 1 (Long Papers)}, pages 2227--2237.

\bibitem[{Radford et~al.(2018)Radford, Narasimhan, Salimans, and
  Sutskever}]{radford2018improving}
Alec Radford, Karthik Narasimhan, Tim Salimans, and Ilya Sutskever. 2018.
\newblock \href
  {https://s3-us-west-2.amazonaws.com/openai-assets/research-covers/language-unsupervised/language_understanding_paper.pdf}
  {Improving language understanding by generative pre-training}.

\bibitem[{Shang et~al.(2015)Shang, Lu, and Li}]{DBLP:conf/acl/ShangLL15}
Lifeng Shang, Zhengdong Lu, and Hang Li. 2015.
\newblock \href {http://aclweb.org/anthology/P/P15/P15-1152.pdf} {Neural
  responding machine for short-text conversation}.
\newblock In \emph{Proceedings of the 53rd Annual Meeting of the Association
  for Computational Linguistics and the 7th International Joint Conference on
  Natural Language Processing of the Asian Federation of Natural Language
  Processing, {ACL} 2015, July 26-31, 2015, Beijing, China, Volume 1: Long
  Papers}, pages 1577--1586.

\bibitem[{Shao et~al.(2017)Shao, Gouws, Britz, Goldie, Strope, and
  Kurzweil}]{DBLP:conf/emnlp/ShaoGBGSK17}
Yuanlong Shao, Stephan Gouws, Denny Britz, Anna Goldie, Brian Strope, and Ray
  Kurzweil. 2017.
\newblock \href {https://aclanthology.info/papers/D17-1235/d17-1235}
  {Generating high-quality and informative conversation responses with
  sequence-to-sequence models}.
\newblock In \emph{Proceedings of the 2017 Conference on Empirical Methods in
  Natural Language Processing, {EMNLP} 2017, Copenhagen, Denmark, September
  9-11, 2017}, pages 2210--2219.

\bibitem[{Sordoni et~al.(2015)Sordoni, Galley, Auli, Brockett, Ji, Mitchell,
  Nie, Gao, and Dolan}]{DBLP:conf/naacl/SordoniGABJMNGD15}
Alessandro Sordoni, Michel Galley, Michael Auli, Chris Brockett, Yangfeng Ji,
  Margaret Mitchell, Jian{-}Yun Nie, Jianfeng Gao, and Bill Dolan. 2015.
\newblock \href {http://aclweb.org/anthology/N/N15/N15-1020.pdf} {A neural
  network approach to context-sensitive generation of conversational
  responses}.
\newblock In \emph{{NAACL} {HLT} 2015, The 2015 Conference of the North
  American Chapter of the Association for Computational Linguistics: Human
  Language Technologies, Denver, Colorado, USA, May 31 - June 5, 2015}, pages
  196--205. The Association for Computational Linguistics.

\bibitem[{Su et~al.(2018)Su, Shen, Hu, Li, and Chen}]{DBLP:conf/aaai/SuSHLC18}
Hui Su, Xiaoyu Shen, Pengwei Hu, Wenjie Li, and Yun Chen. 2018.
\newblock \href
  {https://www.aaai.org/ocs/index.php/AAAI/AAAI18/paper/view/16508} {Dialogue
  generation with {GAN}}.
\newblock In  \cite{DBLP:conf/aaai/2018}, pages 8163--8164.

\bibitem[{Tao et~al.(2018)Tao, Mou, Zhao, and Yan}]{DBLP:conf/aaai/TaoMZY18}
Chongyang Tao, Lili Mou, Dongyan Zhao, and Rui Yan. 2018.
\newblock \href
  {https://www.aaai.org/ocs/index.php/AAAI/AAAI18/paper/view/16179} {{RUBER:}
  an unsupervised method for automatic evaluation of open-domain dialog
  systems}.
\newblock In  \cite{DBLP:conf/aaai/2018}, pages 722--729.

\end{thebibliography}
\bibliographystyle{acl_natbib}
\end{document}